\documentclass[10pt,twocolumn,letterpaper]{article}

\usepackage{iccv}
\usepackage{times}
\usepackage{epsfig}
\usepackage{graphicx}
\usepackage{amsmath}
\usepackage{amssymb}
\usepackage{epstopdf}
\iccvfinalcopy % *** Uncomment this line for the final submission

\def\iccvPaperID{5413} % *** Enter the ICCV Paper ID here

\begin{document}

%%%%%%%%% TITLE
\title{View Confusion Feature Learning for Person Re-identification}

\author{Fangyi Liu
% For a paper whose authors are all at the same institution,
% omit the following lines up until the closing ``}''.
% Additional authors and addresses can be added with ``\and'',
% just like the second author.
% To save space, use either the email address or home page, not both
\qquad Lei Zhang\thanks{ Corresponding author}\\
School of Microelectronics and Communication Engineering, Chongqing University\\
Shazheng street NO.174, Shapingba District, Chongqing 400044, China\\
{\tt\small \{FangyiLiu, LeiZhang\}@cqu.edu.cn}
}

\maketitle
\pagestyle{empty}  % no page number for the second and the later pages
\thispagestyle{empty} % no page number for the first page

%%%%%%%%% ABSTRACT
\begin{abstract}
  Person re-identification is an important task in video surveillance that aims to associate people across camera views at different locations and time. View variability is always a challenging problem seriously degrading person re-identification performance. Most of the existing methods either focus on how to learn view invariant feature or how to combine view-wise features. In this paper, we mainly focus on how to learn view-invariant features by getting rid of view specific information through a view confusion learning mechanism. Specifically, we propose an end-to-end trainable framework, called View Confusion Feature Learning (VCFL), for person Re-ID across cameras. To the best of our knowledge, VCFL is originally proposed to learn view-invariant identity-wise features, and it is a kind of combination of view-generic and view-specific methods. Classifiers and feature centers are utilized to achieve view confusion. Furthermore, we extract sift-guided features by using bag-of-words model to help supervise the training of deep networks and enhance the view invariance of features. In experiments, our approach is validated on three benchmark datasets including CUHK01, CUHK03, and MARKET1501, which show the superiority of the proposed method over several state-of-the-art approaches.
\end{abstract}

\section{Introduction}

\begin{figure}[t]
\begin{center}
   \includegraphics[width=0.7\linewidth]{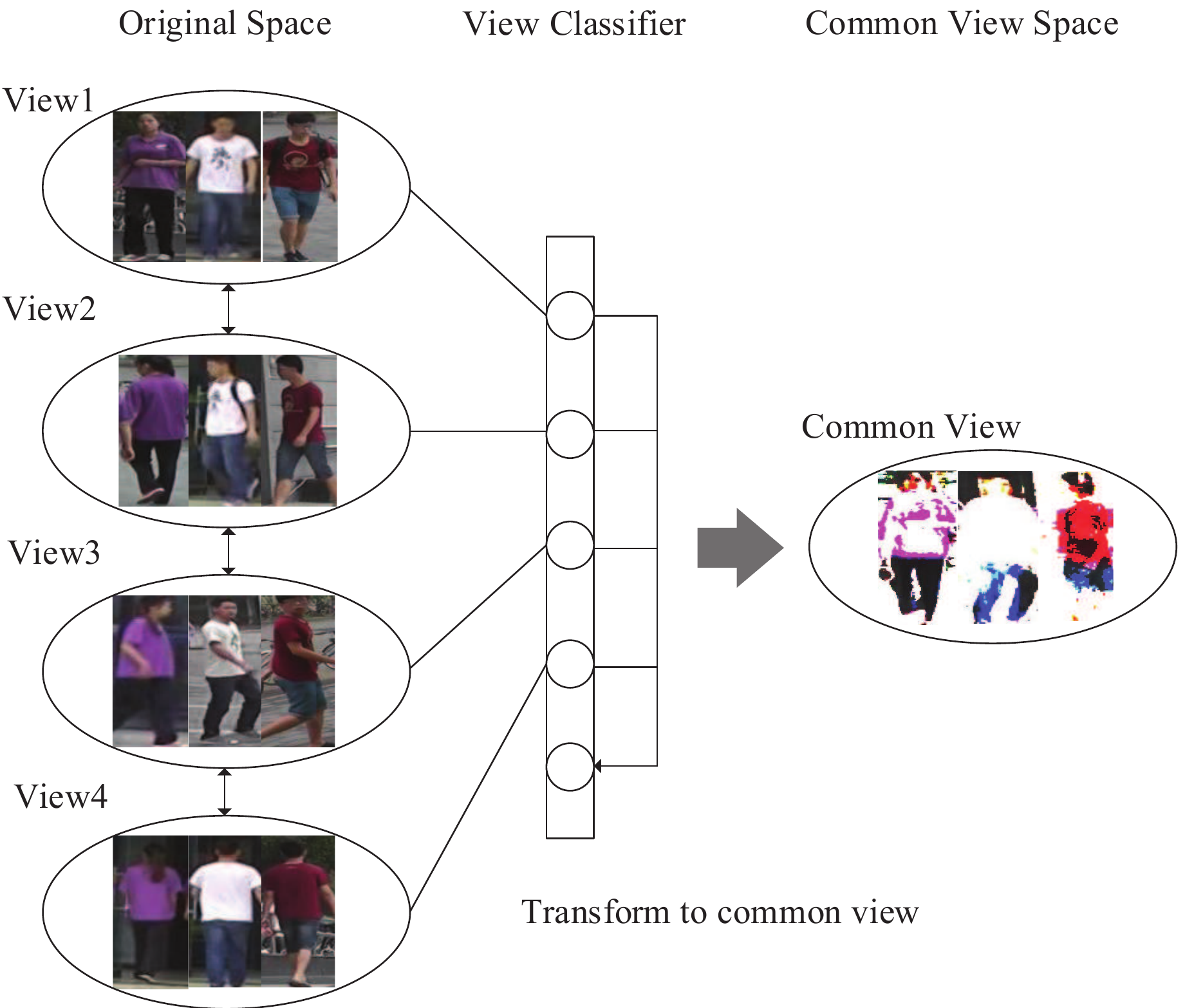}
\end{center}
   \caption{\small Dissimilarity is caused by view changes, and our goal is to achieve view confusion. View confusion is achieved by drawing the distance of each view close and expecting the view classifier to identify each view to common view. Note that the common view is obtained by the average of the 4 views. }
\label{fig1}
\end{figure}

Person re-identification (ReID) is widely applied in many situations such as long-term multi-camera tracking and forensic search. However, due to the problem of non-overlapping area between different camera views, re-identifying pedestrians using appearance features and analyzing their activities across cameras in time and space cues become rather difficult. Results from the variability of camera view, inter-similarity under the same camera becomes more significant than intra-similarity under different cameras. Just as is shown in Figure \ref{fig1}, we aim to solve cross-view problems through view confusion mechanism.

In order to mitigate the influence caused by view variability, existing methods mainly focus on learning robust and discriminative representations \cite{KostingerHirzerWohlhartEtAl2012,LiaoLi2015}, or robust similarity match metrics \cite{ZhangXiangGong2016,HermansBeyerLeibe2017,LiaoLi2015,ShenLinYanEtAl2015} in a supervised manner. Recently, deep learning has gained much attention for learning deep features and metrics in an end-to-end network, and achieves promising results in re-id tasks. Powerful image features should be invariant to variations in illumination, image quality, and especially viewpoint. Many hand-crafted feature types have been used for re-identification, e.g. color, textures, edges and shape, but the discrimination is unsatisfactory. Although deeply learned features have been proved to be powerful in Re-ID tasks, the deep representation is still easily wrapped with view changes. We mainly focus on methods solving cross view changes in deep neural network, which are usually achieved by either designing view-generic models or designing view-specific models with camera view information. View-generic models aim to learn view-invariant features without taking the view information (e.g. pose labels) into consideration, however, they may still suffer from feature distortion caused by camera view variations. The reason is that different views have different impacts on feature extraction, and we can not use only one model to extract features that are invariant to all views. View-specific models aim to utilize camera view information to help cross-view data adaptation and learn view-specific features, however, these features are often limited compared with view-shared features because they are only suitable in specific views. Thus, in this paper, we propose to learn view-invariant features by combining view-generic models and view-specific models, so that our method can be invariant to feature distortion caused by camera view changes.

\begin{figure*}[ht]
\begin{center}
%\fbox{\rule{0pt}{2in} \rule{0.9\linewidth}{0pt}}
   \includegraphics[width=0.8\linewidth]{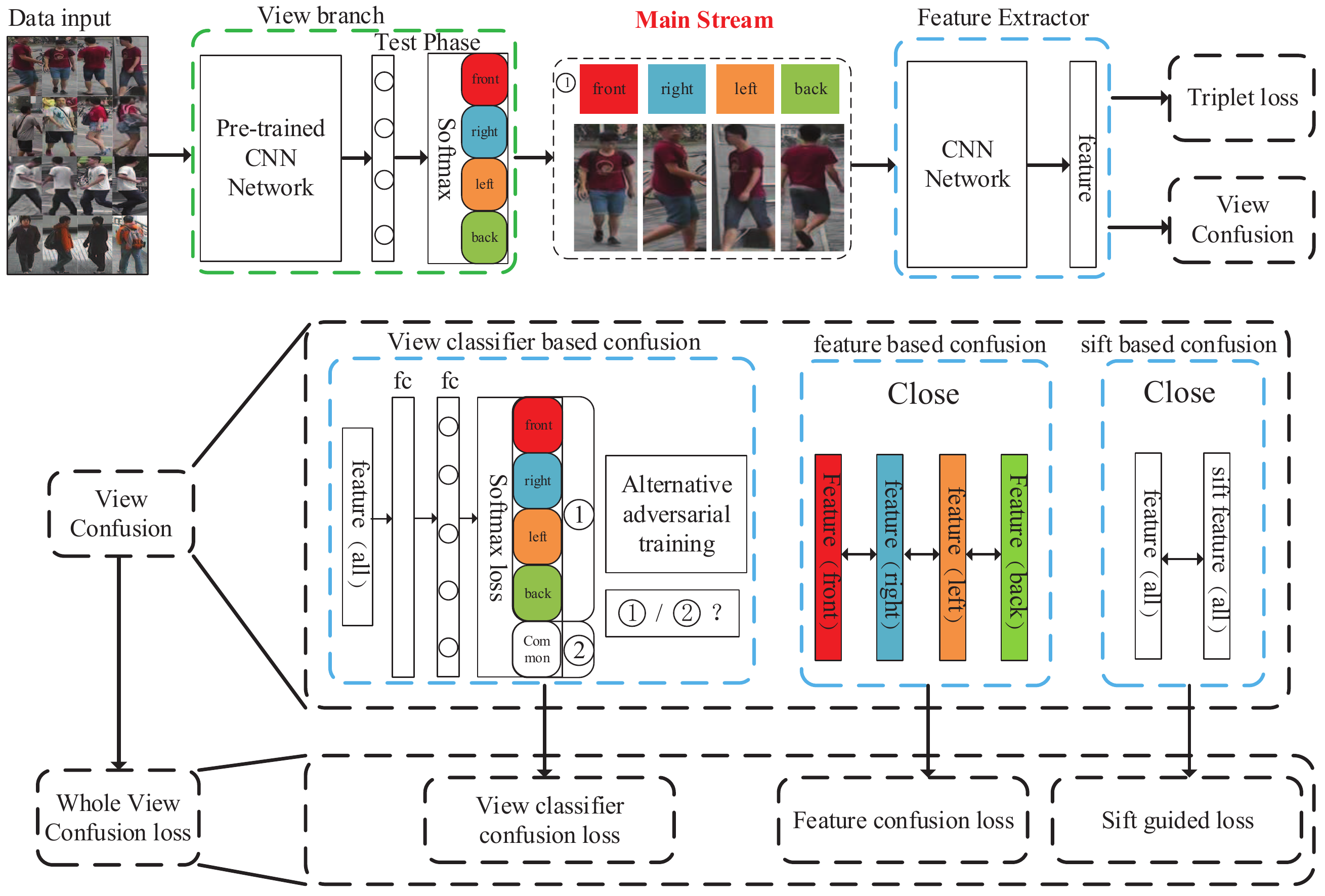}
\end{center}
   \caption{\small Illustration of the proposed VCFL. View branch is pre-trained in RAP dataset \cite{LiZhangChenEtAl2016} for predicting the view information. Our goal is to find the common view features with different view information. View confusion consists of view classifier based confusion, feature based confusion and sift based confusion and it's combined with feature extractor. The whole view confusion loss consists of the loss of three parts as well. The adversarial training ensure that the feature distributions over different views are made similar (i.e. the view classifier can not recognize the view angles of a person), thus resulting in the view-independent features.}
\label{fig2}
%\label{fig:onecol}
\end{figure*}

To achieve view confusion, we consider to propose our method in three aspects: classifier based confusion, feature based confusion and sifted guided confusion. Taking advantages of adversarial thoughts, features are confused to be view-invariant via the iterative training between feature extractor and view classifier. Also, view confusion can be achieved by making features with same label close to feature centers. Further, considering the good interpretability of hand-crafted features (e.g. sift), we propose to take the scale-invariant feature transform (SIFT) \cite{Lowe2004} into account for guiding the learning of the deep feature network. SIFT has the following advantages: first, it is a local feature descriptor of the image, that is invariant to view-changes; second, its distinctiveness is good and informative, that is suitable for rapid and accurate matching in mass feature database; the last but not least, its extensibility can be easily combined with other forms of feature vectors. There are many methods based on SIFT feature, which mostly rely on the bag of word (BOW) methods, before deep learning sprung out. In this paper, we pay more attention to the view-independence of SIFT features, and propose SIFT guided features for better improving the robustness of features.

The contributions of this paper are summarized as follows:
(1) We propose a VCFL approach for learning view-invariant features by using the view confusion learning mechanism. (2) In VCFL, we integrate the SIFT guidance strategy for further improving the view indenpendence of the deep features. (3) Extensive experiments verify the superiority of the proposed VCFL over several state-of-the-art models. \\

\section{Related Work}
%-------------------------------------------------------------------------
\subsection{Person Re-identification}

Existing methods solve person re-identification challenges mainly through two aspects: (1) learning discriminative features which are robust to illumination, poses, view variation and so on. (2) learning similarity metrics which are used to predict whether two images describe the same person. It is well known that discriminative features are important in recognizing. \

However, in order to solve different challenges, these methods try to learn robust features in different ways. In order to solve the problem of pose changes and various human spatial distributions in the person bounding box. Zhao \etal. \cite{ZhaoLiZhuangEtAl2017} propose a simple yet effective human part-aligned representation for handling the body part misalignment problem. Zhang \etal. \cite{ZhangXingWeilaiEtAl2017} propose a novel method called Aligned Re-ID that extracts a global feature which is jointly learned with local features. In order to take advantages of body structure, Zhao \etal. \cite{ZhaoTianSunEtAl2018} propose Spindle Net, based on human body region guided multi-stage feature decomposition and tree-structured competitive feature fusion, in which human body structure information is considered in a CNN framework to facilitate feature learning. Li \etal. \cite{LiChenZhangEtAl2017} design a Multi-Scale Context-Aware Network (MSCAN) to learn powerful features over full body and body parts, it uses attention methods to learn meaningful body parts rather than uses predefined parts which may not be appropriate. \

For learning similarity metrics, most methods propose to solve person re-id problems as ranking problems. Hermans \etal. \cite{HermansBeyerLeibe2017} propose to use a variant of the triplet loss to perform end-to-end deep metric learning, providing guidance for triplet loss training. Chen \etal. \cite{ChenChenZhangEtAl2017} design a quadruplet loss, which can lead to the model output with a larger inter-class variation and a smaller intra-class variation compared to the triplet loss. With the improvements of triplet loss, many end to end framework can gain good performance.

\subsection{Cross View Feature Learning}
It's vital to learn discriminative features which are robust to view variation for person re-id problems. Most of methods for solving cross view challenge can be roughly divided into view-generic methods and view-specific methods. For example, Yu \etal. \cite{YuWuZheng2017} ignored the view information and tried to find a shared space where view-specific bias is alleviated. Feng \etal. \cite{FengLaiXie2018} proposed a deep neural network-based framework which utilizes view information in the feature extraction stage to learn a view-specific network for each camera view with a cross-view Euclidean constraint (CV-EC) and a cross-view center loss (CV-CL). Our proposed VCFL method is a kind of combination of view-generic and view-specific methods. To be specific, view confusion mechanism can utilize view information to remove the impact caused by view variation, such that the model is robust to specific views.

\subsection{Sift Based Methods}
In non-deep learning era, traditional person re-identification methods usually extract low-level features using hand-crafted visual feature descriptors (e.g. SIFT, HOG, etc.). Then the visual retrieval community has witnessed the prominence of the bag-of-words (BoW) \cite{SivicZisserman2003} model for over a decade, during which many algorithms were proposed. The SIFT-based methods for image classification mostly rely on the BoVW model \cite{CsurkaDanceFanEtAl2004}. In this paper, we take advantage of the view independence of sift features to find the view invariant regions in each image to guide the learning of deep model. \\

\subsection{Domain Adaptation}
 Many person re-identification methods attempt to solve cross view problems using domain adaptation methods, because each view can be regarded as an independent domain. Zhong \etal. \cite{ZhongZhengZhengEtAl2018} address cross-view problem by learning a camera-invariant descriptor subspace, it's a kind of camera-style adaptation. Deng \etal. \cite{DengZhengYeEtAl2018} use domain adaptation methods to achieve image translation while maintaining discriminative cues contained in its ID label. It's no wonder that domain adaptation methods are beneficial for solving distribution difference problem in person re-id field. In this paper, the work \cite{GaninLempitsky2015} inspires us a new way to solve view variance, for which we can learn discriminative features for person re-identification task on the main domain and learn invariant features with respect to the shift between the views. To be specific, we introduce the concept of `confusion'. Actually, the view confusion can be understood as view-agnostic, so that the features of a subject from different views can be view-agnostic.

\section{Our Approach}
Person re-id is a task to find the same person across cameras. The challenge is that images of same person taken under different cameras may differ more than images of different person taken under same camera. Recently, most methods treat re-id as a ranking problem, which means the distance between the images of same identities should be closer than those with different identities. Our approach aims at learning features that are robust to view changes. Comparing with other models, view-generic models ignore view information while view-specific models are limited to views. There are also many transfer learning methods trying to turn other view information into \textit{front}. However, it is not hard to find that the transformation matrix may not be suitable to all views, for example, the transformation matrix for \textit{back} to \textit{front} may not be suitable for \textit{left} to \textit{front}. The same problem also exists in view-generic models when trying to learn view-invariant features through only one model. We assume that there must be some common parts (average images) between images with these four kinds of view information, and we call the common parts as common view in later work. It is no wonder that it will be more suitable to transform all views into common view, and the extracted features in common view must be view-invariant. Our approach takes advantage of domain adaptation methods and proposes the concept of `view confusion', which means to get rid of the influence of specific views in feature aspect. Our view confusion is achieved by three parts: classifier based confusion, feature based confusion and sift based confusion. In this section, we will describe the view confusion mechanism based on adversarial idea, SIFT guided feature loss, and the feature learning network.
\subsection{Feature Learning}
Feature leaning has always been an important part in solving person re-id problems, which is beneficial for latter feature matching. Person re-identification tasks are similar to image retrieval in some aspects, for which many methods treat re-id tasks as ranking problems. Our goal is to learn a network which maps images with same id to similar features and map those with different id to different features. To achieve this, we propose to use triplet loss just as \cite{HermansBeyerLeibe2017}. The basic architecture can either be googlenet \cite{SzegedyLiuJiaEtAl2015} or resnet \cite{HeZhangRenEtAl2016}. Triplet loss is proposed to improve the intra-person similarity and inter-person dissimilarity and it's the main loss of our basic network. According to the hard examples mining strategy in \cite{HermansBeyerLeibe2017}, we form the training set into a set of triplets, $ \gamma = {(I_{i}, I_{j}, I_{k})}$, where $(I_{i}, I_{j})$ is a positive pair of images with the same identity and $(I_{i}, I_{k})$ is a negative pair of images with different identity. Then, the triplet loss can be formulated:
\begin{equation}\label{1}
\begin{split}
L_{f} = L_{trip}(I_{i},I_{j},I_{k}) = &[d(h(I_{i}),h(I_{j})) \\
&- d(h(I_{i}),h(I_{k})) + m]_{+},
\end{split}
\end{equation}
where $(I_{i},I_{j},I_{k}) \in \gamma $ , $m$ is the margin by which the distance between a negative pair of images is ensured to be greater than that between a positive pair of images, $h(I)$ represents the extracted feature representation for image I.

\subsection{View Information}
We all know that person poses can be roughly divided into four classes: $\{$'front','right','left','back'$\}$. Since this information depends on the camera as well as the person identity, we call it \textbf{v}iew \textbf{i}nformation (VI) in the remainder of this work. In order to learn a better classifier, we need to get accurate view information of each image. However, it is time-consuming and laborious to manually label these view information of images, and therefore we propose a view branch to predict them \cite{SarfrazSchumannEberleEtAl2018}. Consider that the accuracy of view branch predicting view information matters too much in \cite{SarfrazSchumannEberleEtAl2018}, in our network, the view branch is only used to get view information, and we do not require the prediction accuracy to be as high as possible. This is because our goal is to get rid of these specific view information's influence and obtain view-invariant features.
\subsection{Classifier Based Confusion}
For this part, the confusion is achieved by using view classifier, and our goal is view confusion such that the extracted features can be classified into a common view rather than specific views. Specifically, the proposed confusion consists of two parts: feature extractor and view classifier. The feature extractor tries to learn better features that are robust to view changes while the view classifier tries to identify which view the extracted features belong to. More specifically, the classifier tries to classify features into specific views (front, right, left, back) and the feature extractor tries to learn better features which can be classified into a common view by this classifier. Technically, we can achieve view confusion through adversarial learning strategy. The effectiveness of the view confusion is based on the assumption that features become view-invariant when they can not be classified as any specific views. In other words, the gaming between the feature extractor and the view classifier formulates the Classifier based confusion unit.

At training time, in order to obtain view-invariant features, we seek the parameters $\theta_{f}$ of the feature mapping that maximizes the loss of the view classifier by making the feature distributions as similar as possible, while simultaneously seeking the parameters $\theta_{d}$ of the view classifier that minimizes the loss of the view classifier. Based on this idea, we propose to solve the parameters $\hat\theta_{f}$ and $\hat\theta_{d}$ for feature extractor network and view classifier in an adversarial manner. It can be formulated as follows:
\begin{equation}\label{2}
\begin{split}
&L(\theta_{f},\theta_{d}) = L_{f}(\theta_{f})+L_{d}(\theta_{f},\theta_{d}) \\
&\hat\theta_{f} = arg  \min_{\theta_{f}} L(\theta_{f},\hat\theta_{d}) \\
&\hat\theta_{d} = arg  \min_{\theta_{d}} L(\hat\theta_{f},\theta_{d}).
\end{split}
\end{equation}
As is shown in Figure \ref{fig2}, the feature extractor is used to learn more robust features while the view classifier is used to identify view information. We aim to get better features through adversarial training. We assume that if the extracted features can not be classified into any specific view (or can be classified into a common view) by the trained view classifier/discriminator, then we achieve view confusion. The adversarial classifier based confusion is formulated:
\begin{equation}\label{3}
\begin{split}
&\min \sum_{i=1}^{N} L_{f}^{i}(\theta_{f}) + \lambda \sum_{i=1}^{N} L_{d-}^{i}(\theta_{f},\theta_{d}) \\
&\min \lambda \sum_{i=1}^{N} L_{d+}^{i}(\theta_{f},\theta_{d}),
\end{split}
\end{equation}
where $L_{f}$ is the loss for feature learning (e.g. triplet loss), $L_{d}$ is the loss for the view classifier (e.g. softmax cross-entropy loss), while $L_{f}^{i}$ and $L_{d}^{i}$ denote the corresponding loss functions evaluated at the \textit{i}-th training example. $L_{d+}$ supervise the updating of view classifier to train view classifier better while $L_{d-}$ supervise the updating of feature learning network through view classifier's back-propagation. \textit{N} is the number of training samples and $\lambda$ is set as 0.5 in experiments.

\subsection{Feature Based Confusion}
In order to make the extracted features to be more view-invariant, we try to make features of the same person, which have different view information, as similar as possible. The most direct way is to use center loss \cite{WenZhangLiEtAl2016} which forces features to close to the corresponding feature centers. Center loss intends to learn discriminative features by drawing intra-class distances close while increasing inter-class distances. In \cite{FengLaiXie2018}, View Information is also considered when applying center loss to re-id to further improve the performance, however its goal is to make each sample to be close to view specific and the whole center simultaneously. Our method aims to achieve view confusion in feature aspect which means specific view centers should also close to the whole center, center loss can achieve this without adding any extra computation:
\begin{equation}\label{4}
L_{cen} = \frac{1}{2} \sum_{i=1}^{N} \parallel h(I_{i})-h(C_{y_{i}}) \parallel_{2}, \\
\end{equation}
where $h(I)$ represents the visual features, $C_{y_{i}}$ represents the center (average feature) of identity \textit{y} as shown in Fig. \ref{fig2}, and \textit{N} is sample number. We update the network parameter $\theta$ and the center $C_{y_{i}}$ as follows.
\begin{equation}\label{6}
\begin{split}
&\frac {\partial L_{cen}} {\partial h(I_{i})} = h(I_{i})-h(C_{y_{i}})  \\
&\frac {\partial L_{cen}} {\partial h(C_{y_{i}})} = h(C_{y_{i}})-h(I_{i})  \\
& \theta = \theta - \mu \frac {\partial L_{cen}} {\partial h(I_{i})} \frac {\partial h(I_{i})} {\partial \theta} \\
&C_{y_{i}} = C_{y_{i}}-\alpha \frac {\partial L_{cen}} {\partial h(C_{y_{i}})}.
\end{split}
\end{equation}
where $\mu$ and $\alpha$ represent the learning rate for updating network and center respectively.

\subsection{Sift Based Confusion}
We assume that there exists a kind of view confusion which can be achieved by the adaptive combination of deep features and hand-crafted features. Before deep learning becomes popular with its high accuracy, sift feature takes an import part in the long period of using hand-crafted features. It is meaningful if deep features can be with better quality when combing with sift feature. SIFT features can provide local gradient description, we wonder if the combination of sift and deep features can make features to be with similar distribution, which may help deep features to be more robust to view changes. For each image $x_{i}$ in dataset $\{x^{i}\}^{M}_{i=1}$, we extract SIFT features and then turn them into vectors using BOW model, and we call these vectors as sift-bow vectors. Given the assumption that SIFT features are view-independent, the more the deep features are similar to sift-bow vectors, the more view-independent the deep features are. In other words, we use sift-bow vectors as a supervision to help features leaning, then we propose the sift-guided loss:
\begin{equation}\label{7}
L_{sg} = \sum_{i=1}^{n} \parallel f(x_{i})-g(x_{i}) \parallel_{2}, \\
\end{equation}
where $f(x_{i})$ and $g(x_{i})$ denote deep feature of image $x_{i}$ and sift-bow vector, respectively, and \textit{n} is the number of images.
\begin{equation}\label{8}
\frac {\partial L_{sg}}{\partial f(x_{i}) } = 2 \sum_{i=1}^{n} f(x_{i}).  \\
\end{equation}

\subsection{View Confusion}
The whole model consists of the feature learning part and the view confusion mechanism, and the latter confusion mechanism is achieved by the combination of view classifier based confusion, feature based confusion and sift based confusion. With this combination, the whole loss function can be concluded into feature learning loss and view classifier loss. For feature learning,  triplet loss, feature confusion loss and sift-guided loss are included into $L_{f}$ to keep features' discrimination and view-invariance. For view classifier loss, it is achieved by softmax based cross-entropy loss $L_{d}$ . In the whole loss function of $L_{f}$ and $L_{d}$ are shown as Eq.(\ref{9}) and the update of parameter $\theta$ in the whole model is same as in Eq.(\ref{2}).
\begin{equation}\label{9}
\begin{split}
&L_{f} = \lambda_{fc} L_{fc} + \lambda_{sg} L_{sg} + \lambda_{trip} L_{trip}, \\
&L_{d+} = -\sum_{i=1}^{N} \sum_{c=1}^{4} y_{c}^{i} log^{p_{c}^{i}}, L_{d-} = -\sum_{i=1}^{N} \sum_{c=5} y_{c}^{i}log^{p_{c}^{i}}.
\end{split}
\end{equation}
where $y_{c}^{i}$ and $p_{c}^{i}$ represent view information and the softmax probabilities of the \textit{i}-th image, respectively.

\textbf{An alternative optimization approach:}
Inspired by \cite{GoodfellowPouget-AbadieMirzaEtAl2014}, we use the alternative optimization approach. In the view classifier, two view classifier losses with different view information are included. Different views can be regarded as different domains. Minimization of the first view classifier loss ($L_{d+}$) results in a better domain discrimination, while the second view classifier loss ($L_{d-}$) is minimized when the domains are distinct. Stochastic updates for $\theta_{f}$ and $\theta_{d}$ are then defined as:
\begin{equation}\label{8}
\begin{split}
&\theta_{f} \leftarrow \theta_{f} - \mu(\frac{\partial L_{f}^{i}}{\partial \theta_{f}} + \frac{\partial L_{d-}^{i}}{\partial \theta_{f}} ) \\
&\theta_{d} \leftarrow \theta_{d} - \mu(\frac{\partial L_{d+}^{i}}{\partial \theta_{d}}).
\end{split}
\end{equation}

\begin{figure}[t]
\begin{center}
   \includegraphics[width=0.6\linewidth]{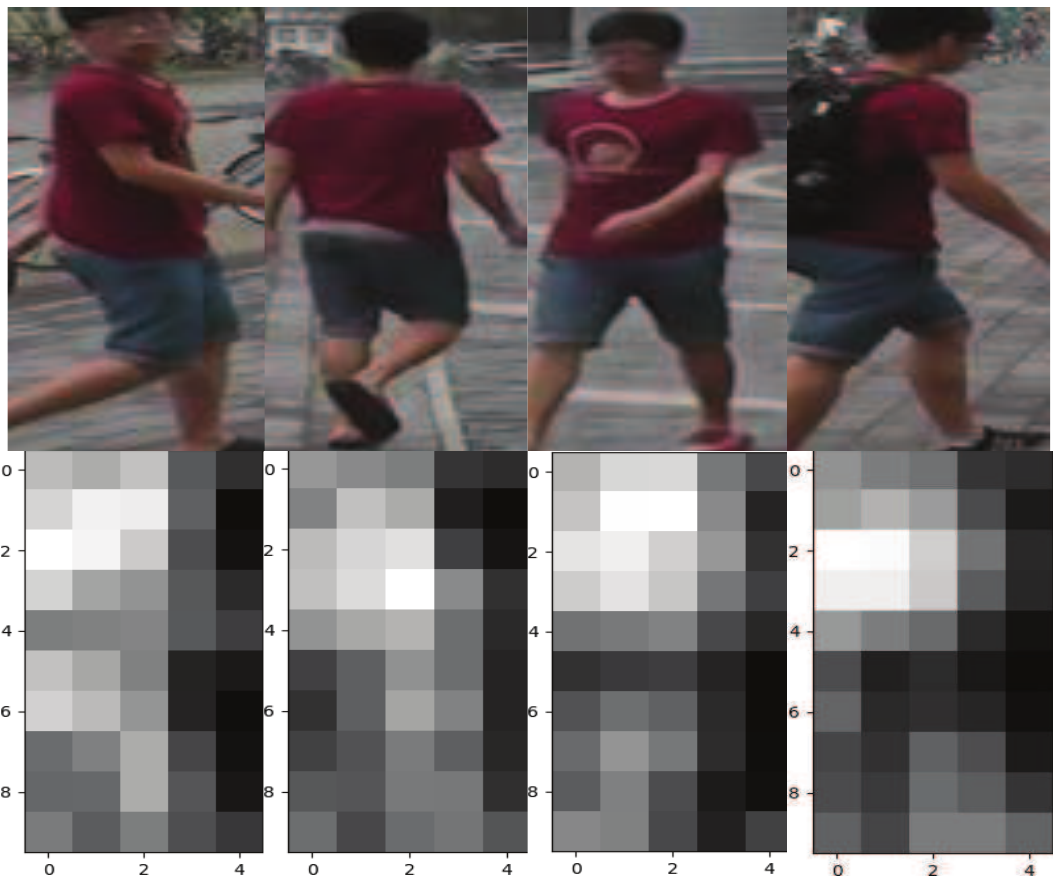}
\end{center}
   \caption{In order to validate the performance of view confusion, we compare the feature map of same id with different view information to show the view invariance of our method.}
\label{fig3}
\end{figure}

\subsection{Implementation Details}
\textbf{Network architecture:}
The network architecture can either be googlenet or resnet. The implementation for googlenet is just as in \cite{ZhaoLiZhuangEtAl2017} that we use a sub-network of the first version of GoogleNet \cite{SzegedyLiuJiaEtAl2015}. It's from the image input to the output of inception 4e, and followed by a 1 $\times$ 1 convolutional layer with the output of 512 channels. Specifically, the person image box is resized to 160 $\times$ 80 as the input, thus the size of the output feature map is 10 $\times$ 5 with 512 channels. However, instead of using part extraction network, we use global average pooling to get final features with 512 channels. The implementation for resnet is just as in \cite{HermansBeyerLeibe2017} that we use resnet \cite{HeZhangRenEtAl2016}. It's from the image input to the output of pooled5 layer, and followed by a fully connected layer with the output of 2048 channels. Specifically, the image input is 384 $\times$ 128.

\textbf{Network training:}
The google network is implemented on Caffe \cite{JiaShelhamerDonahueEtAl2014}. For view confusion part, in order to train a model with view information, we start by fine-tuning the view-predictor branch on the RAP dataset \cite{LiZhangChenEtAl2016}. Next we use the view units to predict view information of target dataset. Specifically, the model trained on RAP dataset is used to fine-tune discriminator which is supposed to identify view information. Similar to GAN \cite{GoodfellowPouget-AbadieMirzaEtAl2014} , the training method of adversarial feature learning is to alternatively train the feature extractor and view classifier. However, we do not need to input noise variables because our goal is to generate more discriminative features instead of synthetic images. The training of feature extractor uses the stochastic gradient descent algorithm (SGD) while that of view classifier uses adaptive moment estimation algorithm (ADAM). In details, we first train the feature extractor to get the initial features, which are then feed into the view classifier. Then we fix $\theta_{f}$ and begin to update $\theta_{d}$.
$L_{d-}$ is given the common information while $L_{d+}$ is given the specific view information predicted by view branch.
The feature learning part is initialized using GoogleNet model that is pre-trained on ImageNet. In each iteration, we sample a mini-batch of 300 images, e.g., there are on average 30 identities with each containing 10 images on Market-1501 and CUHK03. Then, we use python model to get sift-bow vectors for each image in each mini-batch. The goal of feature learning is to get more discriminator features from generator, for which we put forward many supervised identity discriminative information in the generator including triplet loss, center loss and sift-guided loss.
For feature exactor, we adopt the initial learning rate, $\mu_{0}$ = 0.001, and divide it by 10 every 20K iterations. The weight decay is 0.0002 and the momentum for gradient update is 0.9. For view classifier, the momentum for gradient update is 0.9 and the updating strategy is shown as follows:
\begin{equation}\label{7}
\mu_{p} = \frac{\mu_{0}}{(1+\alpha p)^{\beta}},
\end{equation}
where \textit{p} is linearly changed from 0 to 1, $\mu_{0}$ = 0.01, $\alpha$ = 10 and $\beta$ = 0.75.

The ResNet network is implemented on Pytorch. The initial parameters and the training strategy follow \cite{HermansBeyerLeibe2017}. For base net, we use the ResNet-50 architecture and the
weights are provided by He \etal. \cite{HeZhangRenEtAl2016}. The initial learning rate is 0.0003, we fix the learning in the first 151 epochs and then decay following exponentially decaying training schedule. For view classifier, The initial learning rate is 0.001. The momentum for gradient update is 0.9 and the updating strategy is the same. The parameters of feature exactor and view classifier update alternatively, thus increasing the difficulties of training.

\section{Experiments}
\subsection{Datasets and Evaluation Protocol}
\textbf{Datasets:}
Market1501 \cite{ZhengShenTianEtAl2015} contains 32,668 images of 1,501 labeled persons of six camera views. There are 751 identities in the training set and 750 identities in the testing set. In the original study on this proposed dataset, the author also uses mAP as the evaluation criteria to test the algorithms. CUHK03 \cite{LiZhaoXiaoEtAl2014} contains 13,164 images of 1,360 identities. It provides bounding boxes detected from deformable part models (DPMs) and manual labeling. CUHK01 \cite{LiZhaoWang2012} contains 971 identities captured from two camera views in the same campus with CUHK03. Each person has two images, each from one camera view. We report the results of the settings: 100 identities for testing. \

\textbf{Evaluation metrics:}
We adopt the widely-used evaluation protocol \cite{LiZhaoXiaoEtAl2014,AhmedJonesMarks2015}. In the matching process, we calculate the similarities between each query and all the gallery images, and then return the ranked list according to the similarities. All the experiments are conducted under the single query setting. The performances are evaluated by using the cumulated matching characteristics (CMC) curves, which is an estimate of the expectation of finding the correct match in the top \textit{n} matches. We also report the mean average precision (mAP) score \cite{ZhengShenTianEtAl2015} over CUHK03 and Market1501.

\begin{figure}[t]
\begin{center}
   \includegraphics[width=0.8\linewidth]{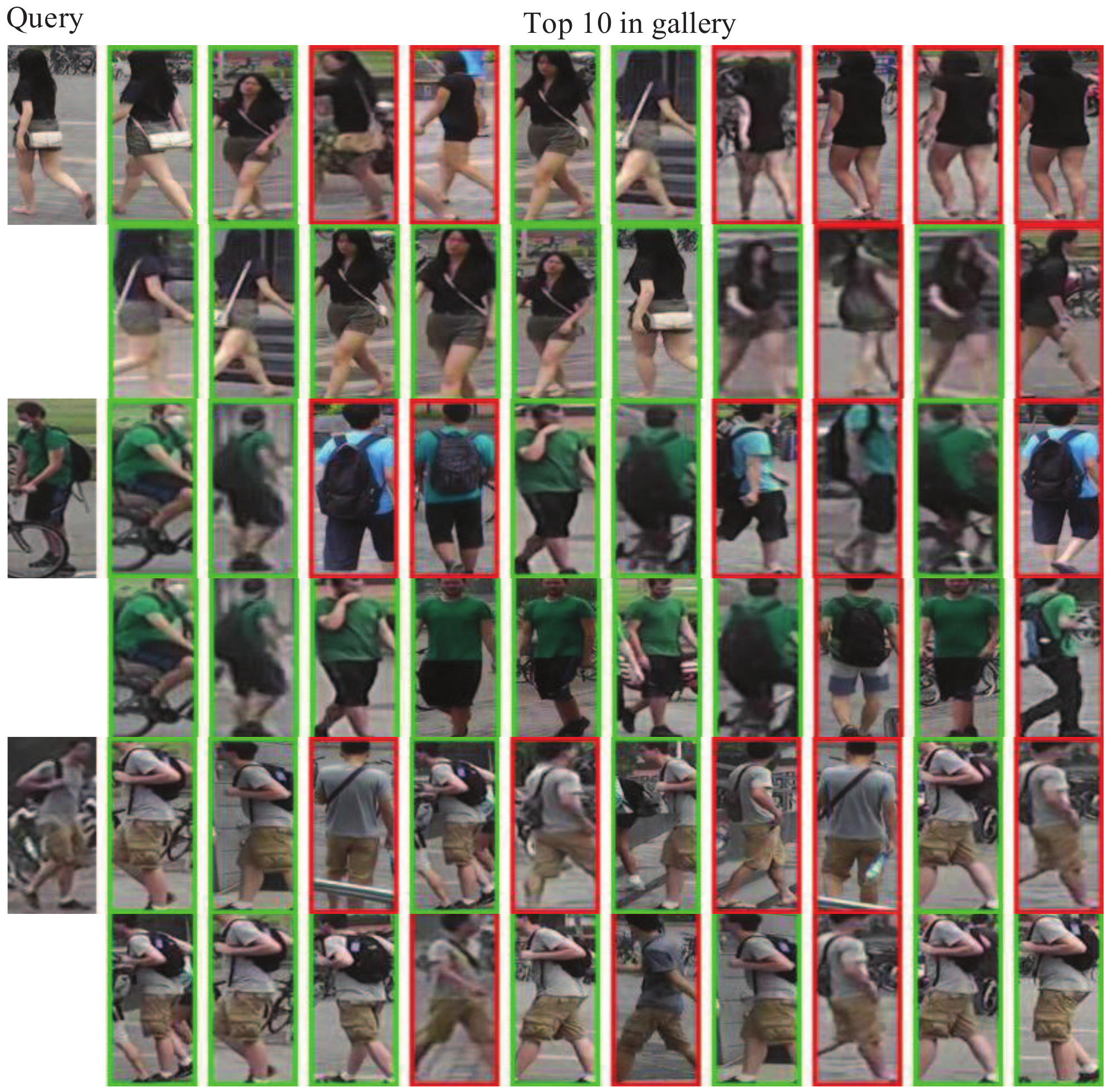}
\end{center}
   \caption{Illustration of the retrieval results on Market 1501. The green rectangle represents a true positive, and the red dash rectangle represents a negative positive. For each samples, the first and second rows show the results for the baseline network representation and our approach representation, respectively. }
\label{fig4}
\end{figure}

\subsection{Comparison With the State-of-the art Methods}

The above experiments have shown the performance of our proposed model. In order to verify the superiority of our method, we compare with the state-of-the-art methods on two popular ReID datasets. We adopt the new train/test protocol with 767 training identities and 700 testing ones which is proposed in paper \cite{ZhongZhengCaoEtAl2017}. The results of cuhk03 (Detected) and maket1501 are shown in Table \ref{Tab3}, and Table \ref{Tab4}, respectively. From the results, we clearly observe the effectiveness and superiority of the proposed method.  Although the accuracy our method is not really high in current re-id field, our model aims to provide new ways for solving re-id problems, thus it can be combined with many other methods to further improve the performance, such as \cite{SunZhengYangEtAl2018}.

\begin{table}
\begin{center}
\caption{The comparison with other methods over CUHK03(D)}
\label{Tab3}
\begin{tabular}{|l|c|c|c|}
\hline
Method & top1 & mAP  \\
\hline\hline
BoW\cite{ZhengShenTianEtAl2015} & 6.36 & 6.39 \\
LOMO\cite{LiaoY.HuLi2015} & 12.8 & 11.5  \\
Resnet50+XQDA\cite{ZhongZhengCaoEtAl2017} & 31.1 & 28.2  \\
Resnet50+XQDA+re-rank\cite{ZhongZhengCaoEtAl2017} & 34.7 & 37.4 \\
SVDNet\cite{SunZhengDengEtAl2017} & 41.5 & 37.3 \\
MultiScale\cite{ChenZhuGong2017} & 40.7 & 37.0 \\
TriNet+Era\cite{ZhongZhengKangEtAl2017} & 55.5 & 50.7 \\
SVDNet+Era\cite{ZhongZhengKangEtAl2017} & 48.7 & 43.5 \\
PCB(UP)\cite{SunZhengYangEtAl2018} & 61.3 & 54.2 \\
PCB(RPP)\cite{SunZhengYangEtAl2018} & 63.7 & 57.5 \\
\hline
baseline & 58.36 & 53.71  \\
VCFL(Ours) & 61.43 & 55.61  \\
VCFL(Ours)+re-rank\cite{ZhongZhengCaoEtAl2017} & \textbf{70.36} & \textbf{70.44}  \\
\hline
\end{tabular}
\end{center}
\end{table}

\begin{table}
\begin{center}
\caption{The comparison with other methods over Market1501}
\label{Tab4}
\begin{tabular}{|l|c|c|c|c|}
\hline
Method & top1 & mAP  \\
\hline\hline
BoW\cite{ZhengShenTianEtAl2015} & 34.4 & 14.09 \\
person net\cite{WuShenHengel2016} & 37.21 & 18.57 \\
WARCA\cite{JoseFleuret2016} & 45.16 & - \\
SCSP\cite{ChenYuanChenEtAl2016} & 51.9 & 26.35 \\
DNS\cite{ZhangXiangGong2016} & 61.02 & 35.68 \\
Gated\cite{VariorHaloiWang2016} & 65.88 & 39.5 \\
Point-to-set\cite{ZhouWangWangEtAl2017} & 70.72 & 44.27 \\
CCAFA\cite{ChenZhuZhengEtAl2018} & 71.8 & 45.5 \\
Consistent-Aware\cite{LinRenLuEtAl2017} & 73.84 & 47.11 \\
Spindle Net\cite{ZhaoTianSunEtAl2018} & 76.9 & - \\
re-ranking\cite{ZhongZhengCaoEtAl2017} & 77.11 & 63.63 \\
GAN\cite{ZhengZhengYang2017} & 78.06 & 56.23 \\
DLPAR\cite{ZhaoLiZhuangEtAl2017} & 81.0 & 63.4 \\
PAN\cite{ZhengZhengYang2018} & 82.8 & 63.4 \\
MultiScale\cite{ChenZhuGong2017} & 88.9 & 73.1 \\
PCB(UP)\cite{SunZhengYangEtAl2018} & 92.3 & 77.4 \\
PCB(RPP)\cite{SunZhengYangEtAl2018} & 93.8 & 81.6 \\
\hline
baseline & 86.58 & 70.91  \\
VCFL(Ours) & 89.25 &  74.48 \\
VCFL(Ours)+re-rank\cite{ZhongZhengCaoEtAl2017} & 90.91 & \textbf{86.67}  \\
\hline
\end{tabular}
\end{center}
\end{table}

\subsection{Analysis of The Proposed Model}
Our experiments are carried on googlenet and resnet to validate our method's performance. The experimental results of googlenet over 3 benchmark datasets are shown in Table \ref{Tab1}. We suppose that the camera view information changes can cause great variation. The proposed view confusion can be integrated into existing methods for further improvement. The training and testing protocol in three dataset is the same as \cite{ZhaoLiZhuangEtAl2017}. In Table \ref{Tab1}, we analysis the performance of our approach and the performance with/without sift guide.
The experimental results of resnet over market1501 are shown in Table \ref{Tab2}. The training and testing protocol of market1501 is the same as \cite{ZhangXingWeilaiEtAl2017}. In Table \ref{Tab2}, we analysis the The performance of our approach and the influence of each part of view confusion, the whole view confusion is achieved by adjusting each part's weight.

\begin{table}
\begin{center}
\caption{The performance of our approach with GoogLenet}
\label{Tab1}
\begin{tabular}{|l|c|c|c|c|}
\hline
Cuhk01-100 & top1 & top5 & top10 & mAP \\
\hline
baseline & 82.3 & 94.6 & 96.4 & - \\
without sift & 88.1 & 96.1 & 97.2 & - \\
whole approach & 86.2 & 94.9 & 97.2 & - \\
\hline\hline

Cuhk03(Detected) & top1 & top5 & top10 & mAP \\
\hline
baseline & 71.7 & 89.2 & 93.1 & 80.2 \\
without sift & 73.29 & 91.36 & 95.93 & 81.51 \\
whole approach & \textbf{76.07} & 93.07 & 96.78 & \textbf{83.70} \\
\hline\hline

Market1501 & top1 & top5 & top10 & mAP \\
\hline
baseline & 75.9 & 89.0 & 92.2 & 55.6 \\
without sift & 76.45 & 89.90 & 92.99 & 56.43 \\
whole approach & \textbf{78.92} & 90.94 & 93.97 & \textbf{58.60}\\
\hline

\end{tabular}
\end{center}
\end{table}

\begin{table}
\begin{center}
\caption{The influence of confusion with ResNet}
\label{Tab2}
\small
\begin{tabular}{|l|c|c|c|c|}
\hline
Market1501 & top1 & top5 & top10 & mAP \\
\hline\hline
baseline & 86.58 & 95.10 & 96.67 & 70.91 \\
classifier based confusion & 85.18 & 94.27 & 96.32 & 69.04 \\
feature based confusion & 87.80 & 95.13 & 96.79 & 73.21 \\
sift based confusion & 88.57 & 95.64 & 97.24 & 74.30 \\
view confusion & 89.25 & 95.61 & 97.18 & 74.48 \\
\hline

\end{tabular}
\end{center}
\end{table}

\textbf{Baseline:} The loss for base net is only the triplet loss.
\emph{GoogLenet:} similar to \cite{ZhaoLiZhuangEtAl2017}, the framework is mainly based on part GoogleNet. However, we do not use the part extraction unit, we use the part GoogleNet instead.
\emph{ResNet:} similar to \cite{HeZhangRenEtAl2016}, the main framework is based on resnet50, and it's pre-trained by weights provided by \cite{HeZhangRenEtAl2016}.

\textbf{View confusion:}
\emph{GoogLenet}: we empirically study how view confusion affects the Re-ID performance. We conduct a experiment over CUHK01 to compare the feature map of the same person with different view information, as is shown in Fig.\ref{fig3}. From the result, we can draw a conclusion that the view confusion can decrease the variations caused by view changes and keep the discrimination, and the feature maps with the same view information are likely to have similar feature distribution after view confusion. We invalidate the performance of the proposed VCFL, and further validate the influence of sift guided loss, cause it's a explorable part of our method. It turns out that the accuracy with sift guided loss increases in cuhk03 and market1501, it may provide us a way to combine hand-crafted feature and deep features to help deep features to be with some good quality of hand-crafted feature.

\emph{ResNet:} we study the influence of each part of view confusion in market1501. As is shown in Table \ref{Tab2}, each part of confusion has contribution in proving the performance. \

\textbf{The influence of view classifier based confusion:} similar to GAN \cite{GoodfellowPouget-AbadieMirzaEtAl2014}, the training of this network is not stable enough, and we adjust the weight smaller when combing with the other two parts. The adversarial learning of the view classifier and the feature extractor matter much in our training stage which means this kind of confusion has a great impact on the final performance. It can provide us a new way to solve cross-view problems however we should train both of the view classifier and the exactor better to produce a positive impact on the final performance. \

\textbf{The influence of feature based confusion:} it is not hard to admit the effectiveness of feature confusion loss, which improves the performance much by drawing different views' feature close. In the training stage, the confusion loss is about $10^{4}$ times of triplet loss, so we just set $\lambda_{fc} = 10^{-4}$ to make the whole loss converge well. Compared with \cite{FengLaiXie2018}, the latter aims to specify all the views and it may suffer the influence caused by the predicted accuracy of view information especially when views increase much. \

\textbf{The influence of sift based confusion:} maybe it is meaningful when applying with deep features, but it also provides us with ways to enhance deep features. In my opinion, sift features are with good locality quality, and this guidance may help deep feature to be with same distribution with this carefully designed feature, thus enhancing features' quality in a way and we set $\lambda_{sg} = 0.1$. \

\textbf{The whole view confusion:} each part of view confusion has different contribution to the final model, and the weights of each part should be carefully selected when combining them together. In this part, we also report our retrial process and ranking result in Figure \ref{fig4}, and it clearly shows that our approach has a good impact on the performance.

\section{Conclusion}
This paper aims to solve view changes in person Re-ID, and prevents the Re-ID system from dropping dramatically due to large variations of camera views and human poses. To improve the performance as well as solve view change problems, we present a sift-guided view confusion adversarial framework for feature learning. Our VCFL is achieved from three aspects: 1) adversarial learning between feature extractor and the view classifier 2) drawing features with same label close to their corresponding centers 3) taking advantages of SIFT's view-independence, it's novel for the combination of hand-crafted features and deep features. Thus, the view-invariant identity-wise features can be learned. In our future work, we'll explore new ways for solving cross-view problems taking advantage of transfer learning and explore the interpretability of deep learning with the help of traditional machine learning methods.

\section*{Acknowledgment}
This work was supported by the National Science Fund of China under Grants (61771079), Chongqing Youth Talent Program, and the Fundamental Research Funds of Chongqing (No. cstc2018jcyjAX0250).

{\small
\bibliographystyle{ieee_fullname}
\bibliography{iccv2019}
}

\end{document}